\documentclass[manuscript,screen,nonacm]{acmart}
\setcopyright{none}
\AtBeginDocument{%
  }

\usepackage{tcolorbox}
\usepackage{algorithm}
\usepackage{amsmath}
\usepackage{algpseudocode}
\usepackage{graphicx}
\usepackage{multirow}
\usepackage{tabularx}
\usepackage{graphicx}
\usepackage{array}

\usepackage{fancyhdr}

\AtBeginDocument{%
  \pagestyle{fancy}
  \fancyhf{}%
  \fancyhead{}
  \fancyfoot[C]{\thepage}
 \fancypagestyle{firstpagestyle}{%
  \fancyhf{}%
}

  \fancypagestyle{plain}{%
    \fancyhf{}%
    \fancyfoot[C]{\thepage}%
  }%
  \fancypagestyle{acmplain}{%
    \fancyhf{}%
    \fancyfoot[C]{\thepage}%
  }%
}

\begin{document}

\title{ContextualSHAP : Enhancing SHAP Explanations Through Contextual Language Generation}

\author{Latifa Dwiyanti}
\email{latifa@stu.kanazawa-u.ac.id}
\orcid{0000-0002-4848-1590}
\affiliation{%
  \institution{Kanazawa University}
  \city{Kanazawa}
  \state{Ishikawa}
  \country{Japan}
}
\affiliation{%
  \institution{Institut Teknologi Bandung}
  \city{Bandung}
  \country{Indonesia}
}

\author{Sergio Ryan Wibisono}
\affiliation{%
  \institution{Institut Teknologi Bandung}
  \city{Bandung}
  \country{Indonesia}}

\author{Hidetaka Nambo}
\affiliation{%
  \institution{Kanazawa University}
  \city{Kanazawa}
  \state{Ishikawa}
  \country{Japan}
}

\renewcommand{\shortauthors}{Dwiyanti et al.}

\setlength{\textfloatsep}{10pt plus 1pt minus 2pt}
\begin{abstract}
 Explainable Artificial Intelligence (XAI) has become an increasingly important area of research, particularly as machine learning models are deployed in high-stakes domains. Among various XAI approaches, SHAP (SHapley Additive exPlanations) has gained prominence due to its ability to provide both global and local explanations across different machine learning models. While SHAP effectively visualizes feature importance, it often lacks contextual explanations that are meaningful for end-users, especially those without technical backgrounds. To address this gap, we propose a Python package that extends SHAP by integrating it with a large language model (LLM), specifically OpenAI’s GPT, to generate contextualized textual explanations. This integration is guided by user-defined parameters (such as feature aliases, descriptions, and additional background) to tailor the explanation to both the model's context and the user's perspective. We hypothesize that this enhancement can improve the perceived understandability of SHAP explanations. To evaluate the effectiveness of the proposed package, we applied it in a healthcare-related case study and conducted user evaluations involving real end-users. The results, based on Likert-scale surveys and follow-up interviews, indicate that the generated explanations were perceived as more understandable and contextually appropriate compared to visual-only outputs. While the findings are preliminary, they suggest that combining visualization with contextualized text may support more user-friendly and trustworthy model explanations.
\end{abstract}

\ccsdesc[500]{Do Not Use This Code~Generate the Correct Terms for Your Paper}
\ccsdesc[300]{Do Not Use This Code~Generate the Correct Terms for Your Paper}
\ccsdesc{Do Not Use This Code~Generate the Correct Terms for Your Paper}
\ccsdesc[100]{Do Not Use This Code~Generate the Correct Terms for Your Paper}

\keywords{XAI, SHAP, Context-Aware Explanation, LLM, OpenAI GPT, Human-Centered AI, Perceived Understandability}


\maketitle

\section{Introduction}
As noted by Molnar \cite{Molnar2022}, interpretability plays a crucial role in encouraging the adoption of machine learning (ML) and artificial intelligence (AI), since transparency helps build trust and supports deployment in high-stakes domains. In line with the goals of AI interpretability, DARPA launched the Explainable AI (XAI) program in 2015 to help users understand, trust and manage AI systems more effectively \cite{Gunning2021}. In 2021, they published a comprehensive summary that summarizes the program's goals, structure, and research progress, further amplifying interest in XAI and inspiring widespread research in the field. 

\begin{figure}[h]
  \centering
  \includegraphics[width=0.6\textwidth]{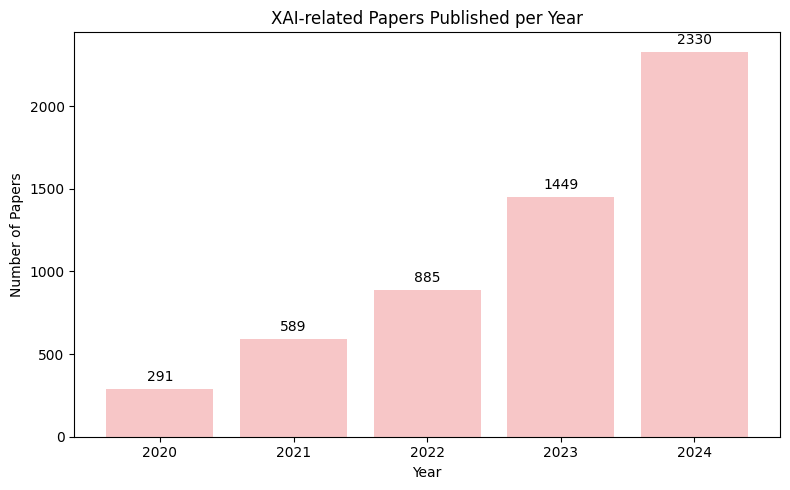}
  \caption{XAI Growing Papers}
  \label{fig:trends}
\end{figure}

Using the open API of the Semantic Scholar platform, which leverages AI through the Semantic Scholar Academic Graph (S2AG) to improve relevance in academic searches \cite{Kinney2023}, the author conducted a query to retrieve articles related to XAI. A total of 5,544 papers were retrieved using three filtering parameters: the keyword "XAI," publication years between 2020 and 2024, and inclusion in the field of Computer Science. The yearly distribution of these publications is shown in Figure \ref{fig:trends}, highlighting the continued growth and increasing interest in this research area.
Currently, there is no universally accepted definition of XAI. The term is often used interchangeably with others such as Interpretable Machine Learning, Reasonable AI, or Understandable AI \cite{Adadi2018}. The goals of XAI can vary widely, as outlined by Arrieta et al. in their comprehensive review \cite{BarredoArrieta2020}, ranging from enhancing trustworthiness, causality, and transferability to promoting informativeness, fairness, interactivity, and awareness of privacy. Among these, the goal most consistently emphasized in all studies is informativeness, the idea that explanations should provide users with sufficient information to support decision making \cite{BarredoArrieta2020}. Arun Rai beautifully captures the essence of XAI by likening it to transforming a 'black box' into a 'glass box', allowing users to understand the rationale behind AI predictions \cite{Rai2020}.

XAI has been applied in a wide range of domains, including medicine, healthcare, law, civil engineering, marketing, education, cybersecurity, transportation, and agriculture \cite{Reddy2023, Yang2023}. However, several challenges persist, such as the trade-off between explainability and model performance, varying human interpretations, the absence of universal standards, biased data leading to unfair outcomes, and the lack of consensus on how to evaluate XAI methods \cite{Reddy2023}. To address human-centric challenges—particularly users with varying abilities to understand explanations—recent efforts have focused on developing context-aware XAI, which generates explanations tailored to both the model and the user's context \cite{Yang2023}. This paper aims to contribute to this direction by introducing a tool for AI researchers to develop models that integrate XAI principles while enhancing the contextual relevance of their explanations.

The structure of this paper is as follows: an introduction, a review of related work, a detailed explanation of our Python library \textit{contextualSHAP}, followed by a case study and evaluation.

\section{Related Works}
\subsection{SHAP (Shapley Additive exPlanation)}
In general, machine learning interpretability methods can be categorized into two types. The first involves models that are inherently interpretable, while the second refers to post hoc interpretability, where explanations are generated after the model has been trained \cite{Molnar2022}. Post hoc interpretability is further divided into two categories: model-agnostic and model-specific approaches. Model-agnostic methods can be applied to any machine learning model, regardless of its internal structure or processing mechanisms, whereas model-specific methods are designed to explain particular types of models \cite{BarredoArrieta2020}.

SHAP is among the most commonly used model-agnostic approaches \cite{Reddy2023,Ma2024,Zhang2020,Linardatos2021}. It enables visualization of the rationale behind model predictions by illustrating feature interactions and importance, both locally and globally, and can be applied to any model or data type \cite{Linardatos2021}. Introduced in 2017, SHAP is based on Lloyd Shapley's 1953 concept of Shapley Values, which quantify a player's average marginal contribution across all possible coalitions \cite{Reddy2023,Lundberg2017}. Equation (1) is the classic Shapley value estimation:

\begin{table}[htbp]
\centering
\begin{minipage}{0.65\textwidth} 
\begin{equation}
\phi_i = \sum_{S \subseteq F \setminus \{i\}} 
\frac{|S|!(|F|-|S|-1)!}{|F|!}
\left[f_{S \cup \{i\}}(x_{S \cup \{i\}}) - f_S(x_S)\right]
\end{equation}
\end{minipage}%
\hfill
\begin{minipage}{0.32\textwidth} 
\scriptsize
\fbox{%
\parbox{\linewidth}{
\textbf{Legend:} \\
$\phi_i$: The Shapley value for feature $i$, representing its overall contribution to the prediction. \\
$F$: The set of all features. \\
$S$: A subset of features not including $i$. \\
$f_{S \cup \{i\}}(x_{S \cup \{i\}})$: The model prediction when feature $i$ is added to subset $S$. \\
$f_S(x_S)$: The model prediction using only the subset $S$. \\
}
}
\end{minipage}
\end{table}

This equation forms the theoretical foundation for SHAP values used in model explanations. However, computing exact Shapley values is computationally expensive, especially for models with many features. To address this, Lundberg et al \cite{Lundberg2020} proposed TreeExplainer, a specialized SHAP implementation for tree-based models such as decision trees, random forests, and gradient-boosted trees. TreeExplainer exploits the internal structure of trees to compute exact Shapley values efficiently, reducing computational complexity to low-order polynomial time \cite{Lundberg2020}.

Algorithm \ref{alg:treeshap} presents a simplified version of the TreeSHAP algorithm, 
which computes SHAP values under *interventional semantics*. 
Specifically, it estimates the interventional expectation 
$\mathbb{E}[f(X) \mid do(X_S = x_S)]$, where $f(X)$ is the model output and $do(X_S = x_S)$ denotes an intervention on the feature subset $X_S$ that breaks dependencies with other features \cite{Lundberg2020}. 
This approach differs from the observational conditional expectation $\mathbb{E}[f(X) \mid X_S = x_S]$, which preserves the natural dependencies between features.

\begin{table}[htbp]
\scriptsize
\centering
\begin{minipage}{0.65\textwidth} 
\begin{algorithm}[H]
\caption{Estimating $E[f(X)\mid do(X_S = x_S)]$}
\label{alg:treeshap}
\begin{algorithmic}[1]
\Procedure{EXPVALUE}{$x,S, \text{tree} = (a,b,t,r,d)$}
  \Procedure{G}{$j$}
    \If{$j$ is internal}
      \State \Return $r_j$
    \Else
      \If{$d_j \in S$}
        \State \Return $G(a_j)$ if $x_{d_j}=t_j$ else $G(b_j)$
      \Else
        \State \Return $G(a_j)\cdot t_{a_j} + G(b_j)\cdot t_{b_j}$
      \EndIf
    \EndIf
  \EndProcedure
  \State \Return $G(1)$
\EndProcedure
\end{algorithmic}
\end{algorithm}
\end{minipage}%
\hfill
\begin{minipage}{0.32\textwidth} 
\fbox{%
\parbox{\linewidth}{
\textbf{Legend:} \\
$x$: The input instance (a complete set of feature values). \\
$S$: The subset of features being conditioned on (i.e., features whose values are known). \\
$a_j, b_j$: The left and right children of node $j$ in the following component. \\
$t_j$: The fraction of data reaching node $j$ that goes into child $a_j$. \\
$r_j$: The output of node $j$ if it is a leaf. \\
$d_j$: The feature index used for splitting at node $j$. \\
}
}
\end{minipage}
\end{table}

\subsection{Generative Language Model}
The effort to enable machines to communicate naturally with humans has been ongoing for decades. Early advancements in natural language processing (NLP) relied heavily on rule-based systems \cite{Kalyan2021}. 

Early systems were resource-intensive and poorly scalable, motivating the development of pretrained language models (PLMs). While PLMs improved performance by leveraging large corpora, they often required task-specific fine-tuning. Large language models (LLMs) addressed this by enabling strong zero-shot and few-shot performance, with GPT-3 marking a key milestone \cite{Brown2020}.

LLMs are based on the Transformer architecture, which uses self-attention to capture long-range dependencies, and are pretrained with self-supervised learning (SSL) through causal language modeling \cite{Kalyan2021, Brown2020, Kalyan2023}. Subsequent variants (InstructGPT, ChatGPT, GPT-4) incorporated reinforcement learning from human feedback (RLHF) to enhance alignment with human intent \cite{Ouyang2022}. The GPT family has since shown substantial advances in natural language generation, with GPT-3 establishing the foundation for human-like text generation \cite{Kalyan2023, Brown2020}.

\subsection{Understandability of Explanation}
As mentioned in the introduction, there is no universally accepted definition of XAI, nor is there a rigid framework for evaluating it. At its core, an explanation generated by an XAI approach serves as an interface between humans and the system, aiming to accurately reflect how a model makes decisions while remaining accessible and understandable to users \cite{Guidotti2018}.

To fulfill XAI’s first goal—describing the reasoning behind a model’s decision—SHAP offers a partial solution by providing both global and local explanations of feature importance. It visualizes how specific features contribute to the model’s output and enables users to see their correlation with the target variable. SHAP operates through a method similar to partial dependence, treating the predictive model as a black box and observing how variations in input features influence predictions \cite{Krause2016}.

However, addressing the second goal—making explanations understandable to users—remains more complex and underexplored. Some studies argue that understandability is mainly intended for developers, who need to interpret and fine-tune models \cite{Miller2017, Miller2019}. But when it comes to end users, there is no standard metric for measuring how well they comprehend model explanations.

To begin, we must consider what “understanding” actually means. This question goes beyond computer science and XAI should be approached as a human-agent interaction problem, intersecting artificial intelligence, social sciences, and human-computer interaction \cite{Miller2019}. The concept of understanding itself can vary. In this paper, we adopt the view that understanding is inseparable from the process of explanation. Drawing on the works of Salmon, Strevens, and Khalifa, we highlight a shared view that scientific understanding fundamentally relies on the ability to construct and assess explanations \cite{Paez2019}.

In the context of XAI, especially when evaluating user comprehension, perceived understandability becomes a crucial concept. It is inherently subjective, depending on each user's individual perception. Sindra Naveed defines perceived understandability as a user’s subjective evaluation of how well they grasp an explanation and how that explanation contributes to their broader understanding of the system's functionality \cite{Naveed2024}. In our previous research, we compared user comprehension across three types of visualizations: (1) a baseline display showing only the input data and classification result, without XAI; (2) a display showing only LIME or SHAP visualizations; and (3) a display integrating the visual explanation with a summary text generated by ChatGPT. Based on a Likert-scale survey measuring perceived understandability, we found that users strongly preferred the combined visual-text explanation. The basic non-XAI output scored the lowest, while the integrated visualization and textual explanation significantly enhanced user comprehension \cite{Dwiyanti2024}. These findings are consistent with the cognitive theory of multimedia learning, which suggests that presentations combining words and images are more effective for understanding than those using only one modality \cite{Mayer2002}.

\section{ContextualSHAP}
Motivated by the potential of combining SHAP visualizations with generative text capabilities offered by ChatGPT, we developed a Python library named ContextualSHAP (https://github.com/latifadwi/contextualshap). Initial experiments using SHAP outputs directly as prompts in ChatGPT revealed limitations in contextual understanding, often resulting in inaccurate or misleading summaries. Therefore, the library was designed to help developers generate more context-aware explanations by offering an easy-to-use, off-the-shelf tool. We hypothesize that enhancing contextual relevance in explanations can lead to improved user comprehension.

\begin{figure}[h]
  \centering
  \includegraphics[width=0.6\textwidth]{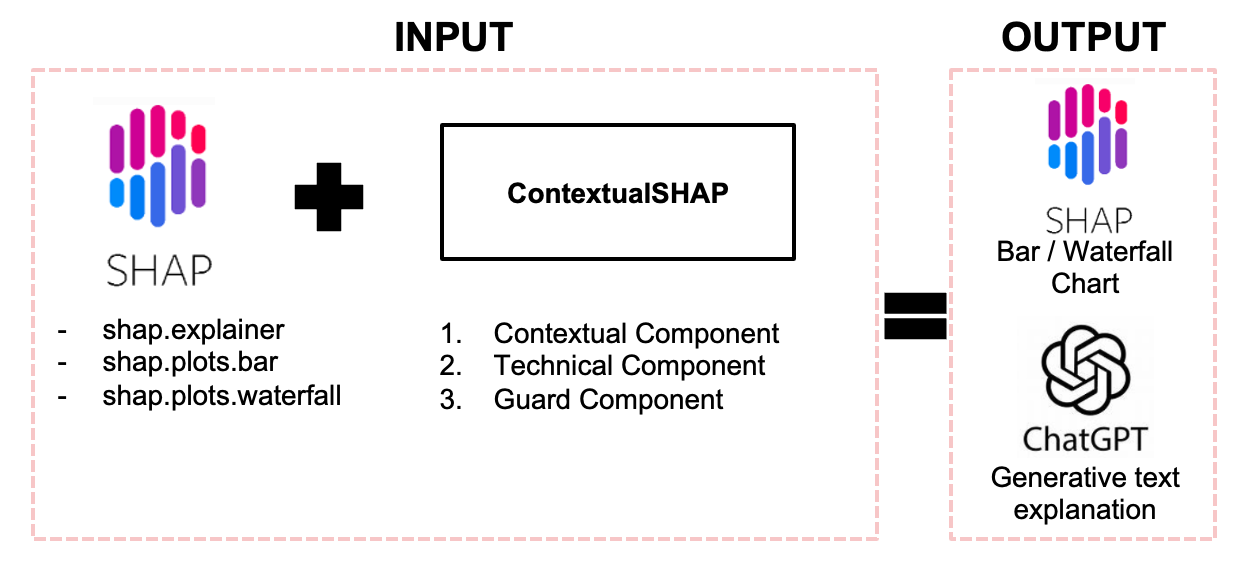}
  \caption{ContextualSHAP}
  \label{fig:contextual}
\end{figure}

The 'contextualSHAP' package extends the functionality of the widely adopted SHAP library, which is primarily used to provide explainability for AI models as part of the broader field of XAI. While the original SHAP library offers robust tools for calculating and visualizing SHAP values, contextualSHAP enhances these capabilities by integrating contextualized natural language explanations through ChatGPT using the OpenAI API. In its initial release, contextualSHAP leverages three key components from the public SHAP API \cite{SHAP2025}: shap\_values, which retrieves shapley values generated by a SHAP explainer; shap.plots.bar,which produces a bar plot representing a set of SHAP values; and shap.plots.waterfall, to visualizes the explanation of a single prediction using a waterfall plot.
These components are wrapped into higher-level utility functions within contextualSHAP to enrich the context provided to ChatGPT, ensuring the resulting explanations are more accurate and user-friendly as depict in Figure \ref{fig:contextual}. The package currently supports the following three core functions.

To provide richer contextual information and enhance the quality of the ChatGPT-generated explanations, the contextualSHAP functions are equipped with five optional parameters. These inputs serve as complementary additions to the SHAP-generated visualizations and help generated explanations more comprehensive, accurate, and user-relevant.

\begin{enumerate}
\item \textbf{Feature renaming} (\texttt{feature\_aliases})  
Feature names inherited from raw datasets or preprocessing are often cryptic or domain-irrelevant. Prior work emphasizes the need for domain-relevant terminology in explanations \cite{Sokol2020}. This parameter enables mapping to meaningful aliases, improving the readability of visual and textual outputs.  

\item \textbf{Feature description} (\texttt{feature\_descriptions})  
Numerical encoding of categorical variables (e.g., sex $\rightarrow$ 0/1) can obscure their meaning. This parameter provides semantic context for such encodings and clarifies feature units or measurement scales (e.g., height in cm vs. ft), ensuring more interpretable explanations.  

\item \textbf{Background context} (\texttt{additional\_background})  
Model explanations are more useful when grounded in application context. This parameter allows users to specify the model’s purpose or task, guiding ChatGPT toward contextually relevant explanations.  

\item \textbf{Language choices} (\texttt{language})  
Explanations are more accessible in a user’s native language \cite{MelbyLervag2014}. This parameter specifies the preferred output language, broadening accessibility across user groups.  

\item \textbf{Reader level} (\texttt{reader})  
Explanations are tailored to audience expertise. Two levels are supported:  
\begin{enumerate}
\item \textit{General}: Non-technical users, with emphasis on intuitive, everyday language.  
\item \textit{Expert}: Domain specialists, emphasizing critical assessment of data, methods, and results.  
\end{enumerate}
\end{enumerate}

Prompts are sent in Markdown format to ChatGPT in the background with enough data to ask ChatGPT to return an explanation that is appropriate for the user. The full text prompts can be found in the package repository, however they contain a few basic aspects needed for the functions provided by the package to work:

\begin{enumerate}
\item A tabular list of feature names, aliases, and descriptions.
These are sent as part of the prompt to let ChatGPT understand the available feature names. When feature aliases/descriptions are not provided, ChatGPT is told to interpret them themselves. ChatGPT has been shown to be able to interpret MedInc as Median Income or Lat/Long as Latitude/Longitude given the proper background context supplied through additional\_background parameter. However, it is still recommended to add per-feature descriptions in addition to background to improve clarity.
\item A limited list of per-feature SHAP values.
Although the user of the package can generate multiple SHAP values for multiple instances of model predictions, only limited number of samples can be used. This is because ChatGPT imposes token limit for each call. SHAP values are presented as a table embedded within the prompt to let ChatGPT understand the numbers without having to read them through images/plots.
\item Plot images. The plot images (waterfall or bar) are also included in the prompt. ChatGPT is also explained whether the image is a waterfall plot of a single instance of prediction (a single SHAP value per feature) or a bar plot containing an average SHAP value for each feature.
\item Additional prompt to format the response. ChatGPT is told to return the response in the exact format to be parsed programmatically.
\end{enumerate}

In addition to the contextual and technical parameters, we embedded a “guard prompt” within the contextualshap package. This prompt acts as a safeguard to ensure that the generative explanation provided by ChatGPT remains within appropriate boundaries and avoids potential misinterpretations or overconfident claims. 

The first purpose of guard prompt is to clarify the nature of predictions. The prompt explicitly reminds the model that it is generating explanations based on predictions from a statistical or machine learning model, not deterministic conclusions. This is especially important in sensitive domains such as healthcare, where outputs from a model (e.g., disease risk predictions) must not be treated as definitive diagnoses. Users are advised to consult with a qualified professional (e.g., a doctor) to interpret any predictive results correctly and responsibly.

The second one, to ensure accurate interpretation of SHAP-specific symbols. During the development and early experimentation with contextualshap, we found that ChatGPT occasionally misinterpreted elements of SHAP plots, particularly the waterfall plot. One recurring issue was confusion between the symbols $E[f(X)]$ and $f(x)$. While $E[f(X)]$ represents the expected prediction value, i.e., the average model output across the entire dataset, the $f(x)$ is the model’s prediction for the specific instance being explained. This kind of confusion can mislead users and compromise the reliability of the explanation. To mitigate this, the guard prompt provides explicit definitions of key symbols and concepts before generating the final explanation.


\section{Implementation and Validation}
To validate the proposed package, we applied it to a real-world case study in the healthcare domain. This domain is particularly appropriate due to its high sensitivity, prediction outcomes can directly influence a patient’s understanding of their health condition. In such scenarios, explanations must go beyond technical accuracy to ensure they are understandable and trustworthy. Moreover, medical datasets often involve specialized terminology, reinforcing the need for contextualized and domain-aware explanations.

\subsection{Implementation : Liver Diseases Classification}
We used the HCV dataset from the UCI Machine Learning Repository. This dataset contains laboratory test results and demographic data from blood donors and patients with Hepatitis C \cite{UCI2025}. The goal of the model is to predict whether an individual has liver disease based on these features.

The raw dataset includes 615 instances and 12 input features, with a target column labeled Category. The Category column includes five values: 0 = Blood Donor (no liver disease), 0s = Suspect Blood Donor (requires further medical assessment), 1 = Hepatitis, 2 = Fibrosis, 3 = Cirrhosis.
We grouped the 1, 2, and 3 labels under a single class called "Liver Disease". Entries labeled 0s were excluded, as they represent uncertain cases requiring expert review. We also removed rows with missing values. After preprocessing, the dataset consisted of 608 complete records. The final class distribution was imbalanced, with 90.4\% labeled as healthy and 9.6\% as liver disease cases. We trained the dataset using XGBoost, which achieved an F1 score of 0.78.

Next, we interpreted the trained model using TreeSHAP, a SHAP-based method tailored for tree-based models. SHAP values were visualized using shap.plots.bar and shap.plots.waterfall. These plots were then passed to ChatGPT with a simple prompt: “Explain this SHAP visualization.” The result depict in Table \ref{tab:without}

\begin{table}[h]
\caption{Image and Explanation without ContextualSHAP}
\label{tab:without}
\scriptsize
\centering
\renewcommand{\arraystretch}{1.5} 
\begin{tabular}{|m{5cm}|m{9cm}|}
\hline 
 \centering \includegraphics[width=0.6\linewidth]{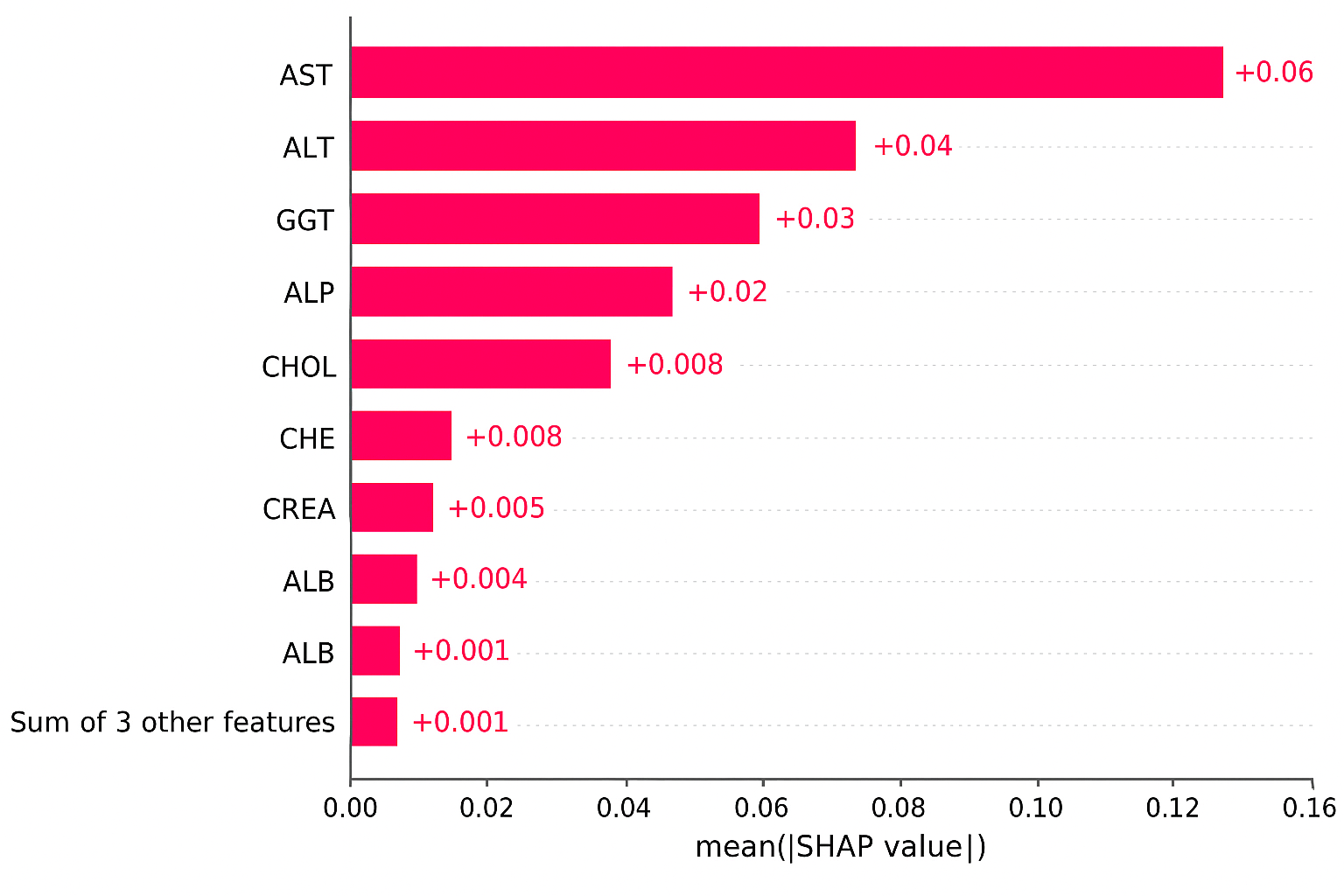} 
& This SHAP plot shows feature importance based on their average impact on the model’s predictions:
\begin{itemize}
    \item AST, ALT, and GGT have the highest influence.
    \item ALP and BIL have moderate influence.
    \item The rest of the features contribute very little.
    \item AST is the top feature, with a mean SHAP value of 0.06.
    \item Some features like ALB have almost no impact.
\end{itemize}
In short: the model relies most on a few top features, especially AST \\ 
\hline
\centering \includegraphics[width=0.6\linewidth]{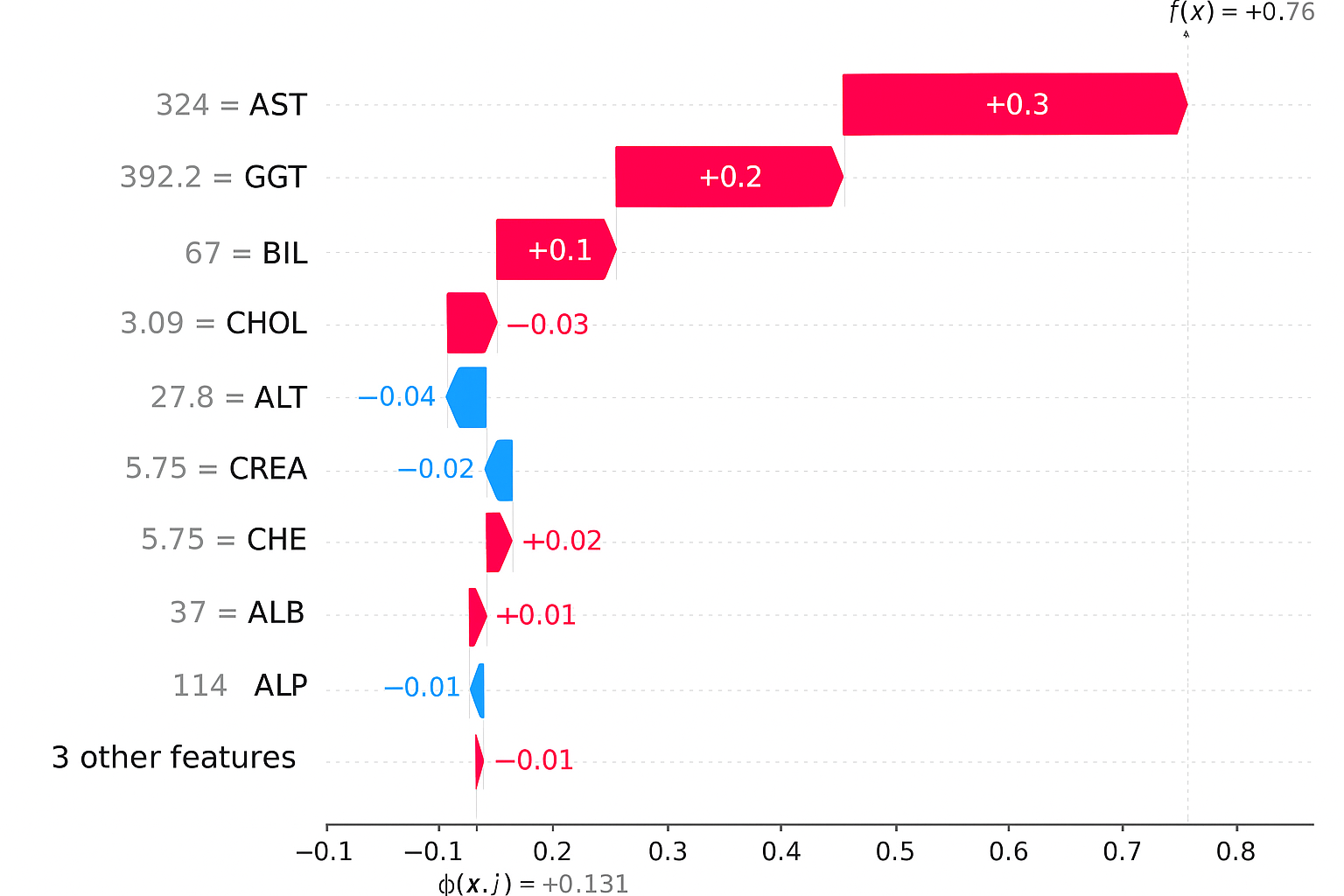} 
& This SHAP plot shows how each feature influenced the model's prediction of 0.76 (starting from a base of 0.131):
\begin{itemize}
    \item Top positive contributors:
    \begin{itemize}
        \item AST (324): +0.3
        \item GGT (392.2): +0.2
        \item BIL (67): +0.1
    \end{itemize}
    \item Top negative contributors:
    \begin{itemize}
        \item ALT (27.8):-0.03
        \item CREA (97.7): -0.02
    \end{itemize}
\end{itemize}
Red means features pushed the prediction higher, blue means they pulled it lower. \\ 
\hline
\end{tabular}
\end{table}

We then applied the contextualSHAP package to the trained model. As a first step, we defined the additional context by specifying the feature\_aliases, feature\_descriptions, and additional\_background parameters. With these contextual inputs configured, we utilized the contextualshap.plots.bar and contextualshap.plots.waterfall functions, setting the language parameter to English, the reader level to general, and using the gpt\-4o model with both explanation and chart display enabled. These functions automatically trigger a call to the OpenAI API to generate the corresponding explanation texts. Using this model, the estimated time per request is approximately 14.9 seconds, with an average cost of about \$0.017 per request. Table \ref{tab:with} depict the plot bar, waterfall bar, and text explanation for this liver disease classification after using the ContextualSHAP library.

\begin{table}[h]
\caption{Image and Explanation with ContextualSHAP}
\label{tab:with}
\scriptsize
\centering
\renewcommand{\arraystretch}{1.5} 
\begin{tabular}{|m{5cm}|m{10cm}|}
\hline 
 \centering \includegraphics[width=0.8\linewidth]{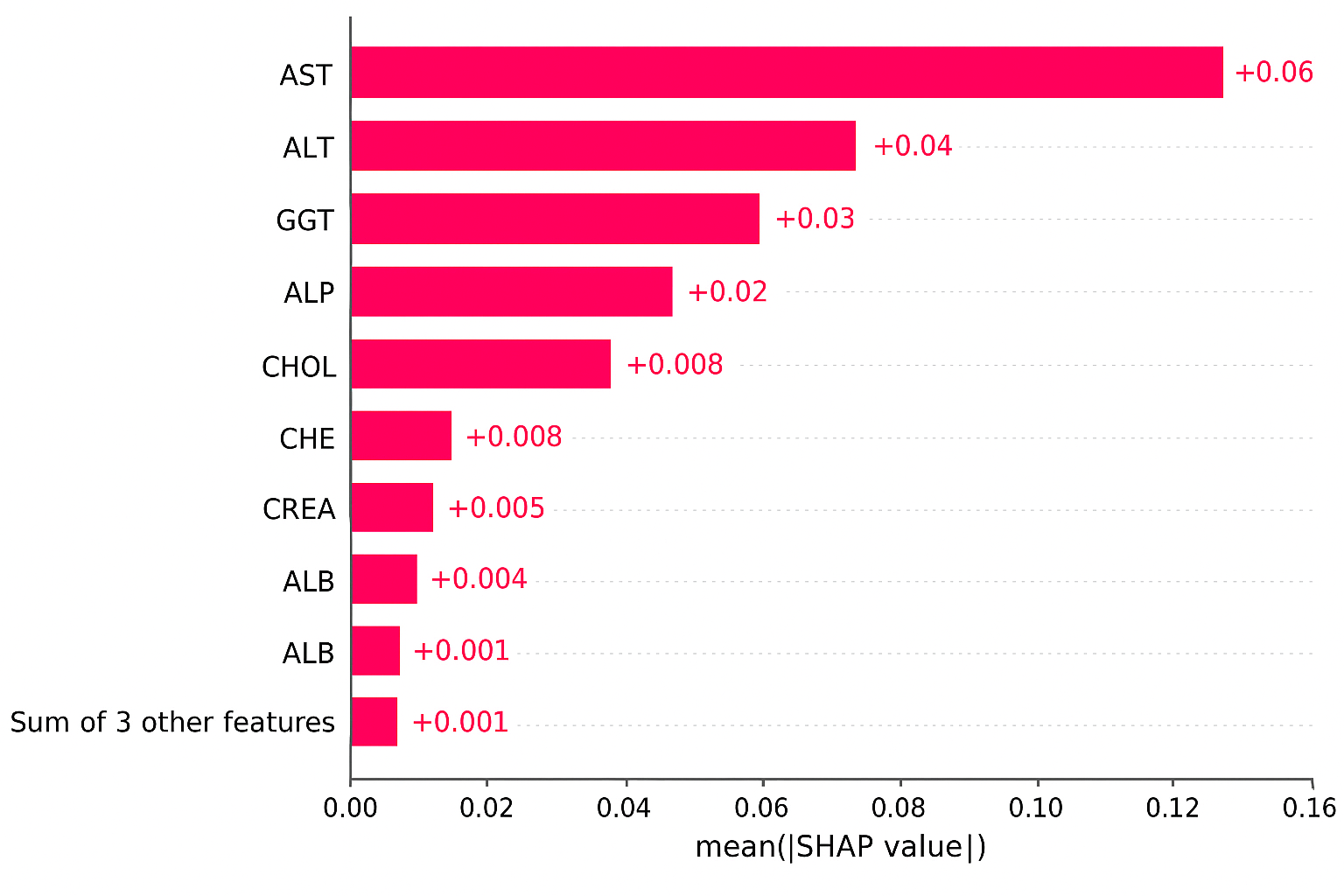} 
& The bar plot you see represents the impact of different features (or patient information) on a model that predicts whether a person has liver disease. These features include various blood test results and demographic information. The longer the bar, the more influence that feature has on the model's decision: 
\begin{enumerate}
    \item AST (Aspartate Aminotransferase): This enzyme is found in the liver and muscles. High levels might indicate liver damage. It has the biggest SHAP value, meaning it's very important in predicting liver disease.
    \item ALT (Alanine Aminotransferase): Like AST, this is another liver enzyme. High levels suggest liver cell damage. It's the second most important factor according to the SHAP value.
    \item GGT (Gamma-Glutamyl Transferase): An enzyme related to the liver and bile ducts. Its SHAP value shows it significantly contributes to the model's prediction, often relating to alcohol use or bile duct issues.
    \item ALP (Alkaline Phosphatase): An enzyme found in various tissues like the liver and bones. Its value suggests moderate importance, usually in liver or bone disorders.
    \item BIL (Bilirubin): It's a pigment from red blood cell breakdown. High levels can indicate liver or bile duct problems. The model finds it moderately significant.
    \item CHOL (Cholesterol): Though known for heart disease risk, its role in the model suggests its levels might have some correlation with liver health.
    \item CHE (Cholinesterase): A liver-produced enzyme, and low levels can indicate liver issues. It appears mildly significant.
    \item CREA (Creatinine): High levels indicate kidney function problems, which indirectly relates to overall health status the model considers.
    \item ALB (Albumin): A protein by the liver. Surprisingly, the model doesn't find it very influential, as shown by its low SHAP value, but it could still be relevant in specific cases.
    \end{enumerate}
    Other Features: Include the rest of the features like Age and Sex, which the model uses but finds less critical than the main ones listed. \\
\hline
\centering \includegraphics[width=0.8\linewidth]{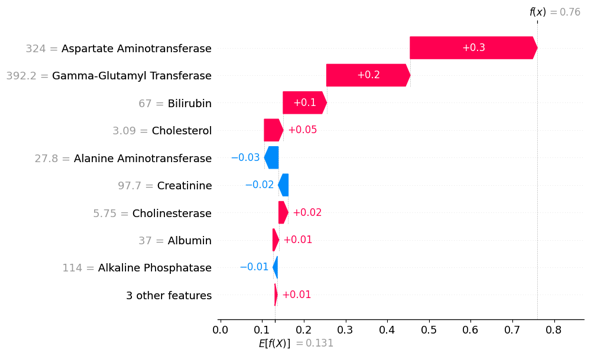} 
& The waterfall plot explains how different features contribute to predicting the likelihood of a patient having liver disease. The prediction starts at an average baseline value and gets adjusted up or down depending on certain factors, called SHAP values.
\begin{enumerate}
    \item AST (Aspartate Aminotransferase): This is a liver enzyme. Here, with a high SHAP value of +0.30, it strongly pushes the prediction towards indicating liver disease. The patient's AST level is 324, much higher than normal, suggesting possible liver damage.
    \item GGT (Gamma-Glutamyl Transferase): Related to bile ducts, this enzyme's SHAP value of +0.20 also significantly pushes the prediction towards liver disease, with a patient's level of 392.2, indicating potential liver or bile duct problems, possibly due to alcohol use.
    \item BIL (Bilirubin): High bilirubin levels can indicate liver problems. With a SHAP value of +0.10, this feature also contributes to a higher chance of liver disease, given the patient's level of 67.
    \item CHOL (Cholesterol): It slightly pushes the prediction towards a liver disease probability with a SHAP value of +0.05. The patient's value is 3.09, which needs further context to interpret fully.
    \item ALT (Alanine Aminotransferase): With a SHAP value of -0.03, this enzyme actually pulls the prediction away from liver disease, indicating the patient's level (27.8) might be in a safer range.
    \item CREA (Creatinine): This helps evaluate kidney function. A SHAP value of -0.02 means it slightly reduces the chance of liver disease. The patient's level is 97.7.
    \item CHE (Cholinesterase): With a slight SHAP value of +0.02, it leans towards liver disease, as a low level can indicate liver issues. The level here is 5.75.
    \item ALB (Albumin): A protein made by the liver, it marginally indicates liver disease with a SHAP value of +0.01 as the patient’s level is 37.
    \item ALP (Alkaline Phosphatase): It has a small SHAP value of -0.01, which lightly pushes the prediction away from liver disease. The patient's level is 114.
\end{enumerate}
Each feature in the waterfall plot shows its influence, positive or negative, on the prediction. Features adding positively increase the likelihood of predicting liver disease, whereas negative ones suggest otherwise. In this case, the model prediction finally settles at 0.76, indicating a relatively high risk of liver disease for the patient.\\ 
\hline
\end{tabular}
\end{table}

\subsection{Validation}

To validate whether the proposed approach improves end-user understandability, we conducted two types of evaluations: a Likert-scale survey and semi-structured deep interviews. These methods were applied to seven participants with different backgrounds, as shown in Table \ref{tab:participant-groups}.

\begin{table}[h]
  \caption{Group Participant Demographics}
  \label{tab:participant-groups}
  \small
  \begin{tabular}{lcc}
    \toprule
    & Laypersons & Experts (Doctors) \\
    \midrule
    Number of participants & 5 & 2 \\
    Age (years) & 24--39 (average 31.8) & 31, 34 \\
    Nationality & Japanese (2), Indonesian (2), Chinese (1) & Indonesian (2) \\
    \bottomrule
  \end{tabular}
\end{table}

In the Likert-scale survey, users were asked to rate their agreement (1 = Strongly Disagree to 5 = Strongly Agree) on five questions measuring perceived understandability across three types of explanations: (i) SHAP-only visualization, (ii) SHAP + non-contextual explanation, (iii) contextualSHAP: SHAP + contextual generative text. The result of the average likert-scale survey depict in Table \ref{tab:user}.

\begin{table}[h]
\small
\centering
\caption{User Perceived Understandability (Average Likert Scores)}
\label{tab:user}
\begin{tabularx}{\textwidth}{X|c c c|c c c}
\hline
\multirow{2}{*}{Questions} & \multicolumn{3}{c|}{Mean Score Group 1} &
\multicolumn{3}{c}{Mean Score Group 2} \\
\cline{2-7}
 & i & ii & iii & i & ii & iii \\
\hline
The explanation helped me understand how the model arrived at its decision. & 1.8 & 2.8 & 3.2 & 3 & 4 & 4 \\
The explanation was clear and easy to comprehend. & 1.6 & 2.6 & 3.2 & 3 & 4 & 4.5 \\
The explanation felt relevant to my knowledge and background. & 1.2 & 1.6 & 2.6 & 4 & 4 & 4.5 \\
The combination of text and visual elements improved my understanding. & NA* & 3.2 & 3.6 & NA* & 3.5 & 4 \\
I feel confident that I understood the model’s reasoning after reviewing the explanation. & 1.6 & 3 & 3.8 & 3 & 4 & 4 \\
\hline
Average & 1.55 & 2.64 & 3.28 & 3.25 & 3.9 & 4.2 \\
\hline
\end{tabularx}

\medskip
\small \textit{Note}: NA indicates the question was not applicable to SHAP-only visualization due to the absence of explanatory text.
\end{table}

Based on the Likert-scale responses and interviews, the layperson group showed the clearest improvement in perceived understandability when moving from SHAP-only visualizations (i) to explanations with text (ii), and further to contextual explanations (iii). While generic explanatory text (ii) already improved clarity compared to raw SHAP outputs, participants reported that contextualized text (iii) was significantly more effective in bridging the gap between technical information and their background knowledge. This progression highlights the importance of contextualization in addition to textual support. Consistent with the cognitive theory of multimedia learning, the combination of visuals and tailored text substantially improved comprehension. Furthermore, some participants emphasized the added value of receiving explanations in their native language, reinforcing the importance of localization for accessibility.

In contrast, the doctor group reported consistently high levels of understandability across all formats. For them, the difference between generic (ii) and contextual explanations (iii) was marginal, as their domain expertise enabled them to interpret SHAP outputs directly. They noted that while textual explanations are not harmful, conciseness and focus are more valuable than elaboration. This indicates that explanation strategies should be user-adaptive, in this case laypersons benefit most from contextualized, localized explanations, whereas experts prefer succinct interpretations.

\section{Result}

This study aimed to improve the perceived clarity of model explanations by combining SHAP visualizations with context-aware natural language outputs generated via a custom Python package, contextualSHAP, which uses OpenAI’s language model. Effectiveness was assessed through a Likert-scale survey and interviews with a layperson and a medical expert, comparing three conditions: (i) SHAP only, (ii) SHAP with generic explanation, and (iii) SHAP with context-aware explanation from contextualSHAP.

The results showed that contextualSHAP consistently improved understandability scores, particularly for the layperson group, who considered the added context essential for interpreting the model’s decisions. Their average scores rose from 1.55 (SHAP-only) and 2.64 (SHAP+generic) to 3.28 (contextualSHAP), demonstrating a clear gain in perceived clarity. Participants also emphasized that receiving explanations in their native language significantly enhanced comprehension, underscoring the role of localization in accessibility. In contrast, the doctor group reported high scores across all conditions, with only marginal differences between generic and contextual explanations (3.25, 3.9, and 4.2, respectively). Their domain expertise allowed them to interpret SHAP outputs without additional context, and they expressed a preference for concise, focused outputs over elaboration.

These findings highlight the importance of user-tailored, context-aware explanations, as different user groups perceive them differently. This paper reports a preliminary experiment, and further studies are needed to confirm that contextualSHAP definitively improves the understandability of machine learning model predictions. Future research should involve larger participant groups across diverse domains to validate the effectiveness of contextual explanations and to further assess the flexibility and impact of contextualSHAP in supporting human-centric AI explainability.


\bibliographystyle{unsrt}
\bibliography{sample-base.bib}

\appendix

\end{document}